\pdfoutput=1
\documentclass[letterpaper, 10 pt, conference]{ieeeconf}  

\IEEEoverridecommandlockouts                              

\usepackage[T1]{fontenc}
\usepackage[latin1]{inputenc}
\usepackage{times}
\usepackage{subfig}
\usepackage{lmodern}
 \usepackage[justification=centering]{caption}

\usepackage{tabularx}
\usepackage{array}
\usepackage{pgfplots}
\usepackage{tikz}
\usepackage{amsmath} 
\usepackage{amssymb}  

\usepackage{graphics} 
\usepackage{epsfig} 
\usepackage{mathptmx} 
\usepackage{epstopdf}
\usepackage{nicefrac}
\usepackage{float}
\usepackage{url}

\usepackage[letterpaper, margin=0.75in]{geometry}

\title{\Large\bf LiDAR point clouds correction acquired from a moving car based on CAN-bus data}
\author{Pierre Merriaux$^{1}$, Yohan Dupuis$^{2}$, R\'{e}mi Boutteau$^{1}$, Pascal Vasseur$^{3}$ and Xavier Savatier$^{1}$
\thanks{$^{1}$Pierre Merriaux, R\'{e}mi Boutteau and Xavier Savatier are with IRSEEM, ESIGELEC, 76800 Saint-Etienne-du-Rouvray, France {\tt\small firstname. lastname@esigelec.fr}}%
\thanks{$^{2}$Yohan Dupuis is with Department of Multimodal Transportation Infrastructure, CEREMA, 76120 Le Grand Quevilly, France {\tt\small yohan.dupuis@cerema.fr}}%
\thanks{$^{3}$Pascal Vasseur is with LITIS Lab, University of Rouen, 76821 Saint-Etienne-du-Rouvray, France {\tt\small pascal.vasseur@univ-rouen.fr}}%
}

\begin{document}

\maketitle
\thispagestyle{empty}
\pagestyle{empty}

\begin{abstract}
In this paper, we investigate the impact of different kind of car trajectories on LiDAR scans. In fact, LiDAR scanning speeds are considerably slower than car speeds introducing distortions. We propose a method to overcome this issue as well as new metrics based on CAN bus data. Our results suggest that the vehicle trajectory should be taken into account when building 3D large-scale maps from a LiDAR mounted on a moving vehicle.
\end{abstract}

\section{INTRODUCTION}
%
%
%
%




Companies involved in the automotive industry, research laboratories, car manufacturers, automotive suppliers, have been aiming at developing more and more automated driving systems. The ultimate goal is represented by fully autonomous cars.

Promising results have been achieved with the use of LiDAR systems. First based on mono-layer LiDAR, recent breakthroughs are now reached with multi-layer LiDAR. The most famous example is the autonomous car developed by Google. Car manufacturers, such as Ford, have also based their autonomous driving solution on multi-layer LiDAR. Multi-layer LiDAR gives 3D information. They allow multi-tasking as they enable to see all the components from a road scene: the road itself, other vehicles, the environment. It allows to detect road markings, necessary for lane-keeping purposes, while monitoring other objects, static or dynamic, in the vicinity of the vehicle. All information can be fed to a decision layer in order to perform path planning and control.

More advanced autonomous applications require to locate the vehicle within its environment and not locally on the road. Localization is challenging in urban environments. In fact, centimeter grade localization is nowadays performed by mixing IMU information and RTK-GPS. These technologies are expensive. Consenquently, another approach is finding the more likely position given a 3D map created before the vehicle travels the mapped area : the localization problem.

3D map are generally build based on SLAM techniques. However, drifting is really important in large-scale environments ans loop-closure might not be available.

LiDAR are mounted on road vehicles. Their positions are obtained from highly accurate IMU and RTK-GPS. Such precise location of LiDAR measurements enable to build good 3D map of the environment. Still, acquiring data from a moving car is more challenging as the car can move at fast speeds contrary to indoor robots. Consequently, the car motion should be taken into account in order to correct the 3D scans. In fact, LiDAR output scans including all the beams once at a time. The beams are often considered as being taken at a single time as in~\cite{Baldwin2012}. The measurement process of a LiDAR makes that each beams within a scans are captured at different times.

Existing approaches rely on either LiDAR data to estimate the car motion or use dedicated sensors. First, SLAM frameworks can be used to estimate the motion performed by the car during the last available scan by matching the last scan with the previous one. Then, the LiDAR location is interpolated within the scan to take into account the LiDAR motion during the scan. For instance, ~\cite{Hong2010} proposed a variant of ICP that estimate the LiDAR velocity by intrinsically correcting the distortion. In~\cite{Zhang2014}, Zhang et al relies on different update rates to mutually estimate the map and the LiDAR odometry. The scan registration is based on keypoint matching. Then, once the odometry is estimated, the previous scans are updated to take into account the LiDAR motion during the scanning process.  Secondly, motion estimation sensors can be used instead of LiDAR-based odometry estimation. In~\cite{Byun2015}, Byun et al use a high-end GPS/INS unit to obtain the 3D odometry of the car. It requires to embed a extra sensor to remove the motion-based distortion within the scan.


In this paper, we investigate the effect of different vehicle movement on LiDAR scan distortions. A method is proposed to correct the scan given the intra-scan vehicle motion estimated solely from CAN bus data. Experiments are carried out in a real environment. The corrected scan are used in a standard SLAM framework that do not include scan distortion removal. Results shows that translations and rotations should be taken into account in order to obtain consistent 3D maps.




\section{METHODOLOGY}

We aim at evaluating the performance of LiDAR-based localization of road vehicles. Consequently, we must use large-scale 3D map of the environment. Such maps are not freely available. Consequently, we must build them from a moving LiDAR mounted on a car. LiDAR scans are distorted by the motion of the vehicle on which the LiDAR is mounted. As we will see later, distortions can be large and reduce the accuracy of the map build from such data. LiDAR scans must be corrected prior to the fusion of the maps acquired at different locations.

There exists two ways to overcome LiDAR scan distortions induced by the motion of the LiDAR while scanning:
\begin{itemize}
\item Increase scanning speed. The larger the scanning speed, the smaller the motion of the LiDAR during a scan, the smaller the distortions. This approach can easily be applied for mobile robots which are generally slower the road vehicles~\cite{Merriaux2015b}.
\item Correct LiDAR scans from motion estimation. Motion can be estimated from sensors such as IMU, odometry sensors or cameras~\cite{zhang2015v},\dots or processing by two consecutive scans \cite{moosmann2011v} to achieve LiDAR-based odometry.
\end{itemize}

\cite{dong2013p} presents different ways to interpolate the LiDAR trajectory when the scanning speed is low or when two scans of a tilting LiDAR are required~\cite{nuchter2007p}.

	\begin{figure}[t]
	\centering
	    {\includegraphics[width=0.8\linewidth, clip=true,trim=2.5cm 17cm 5.5cm 4cm]{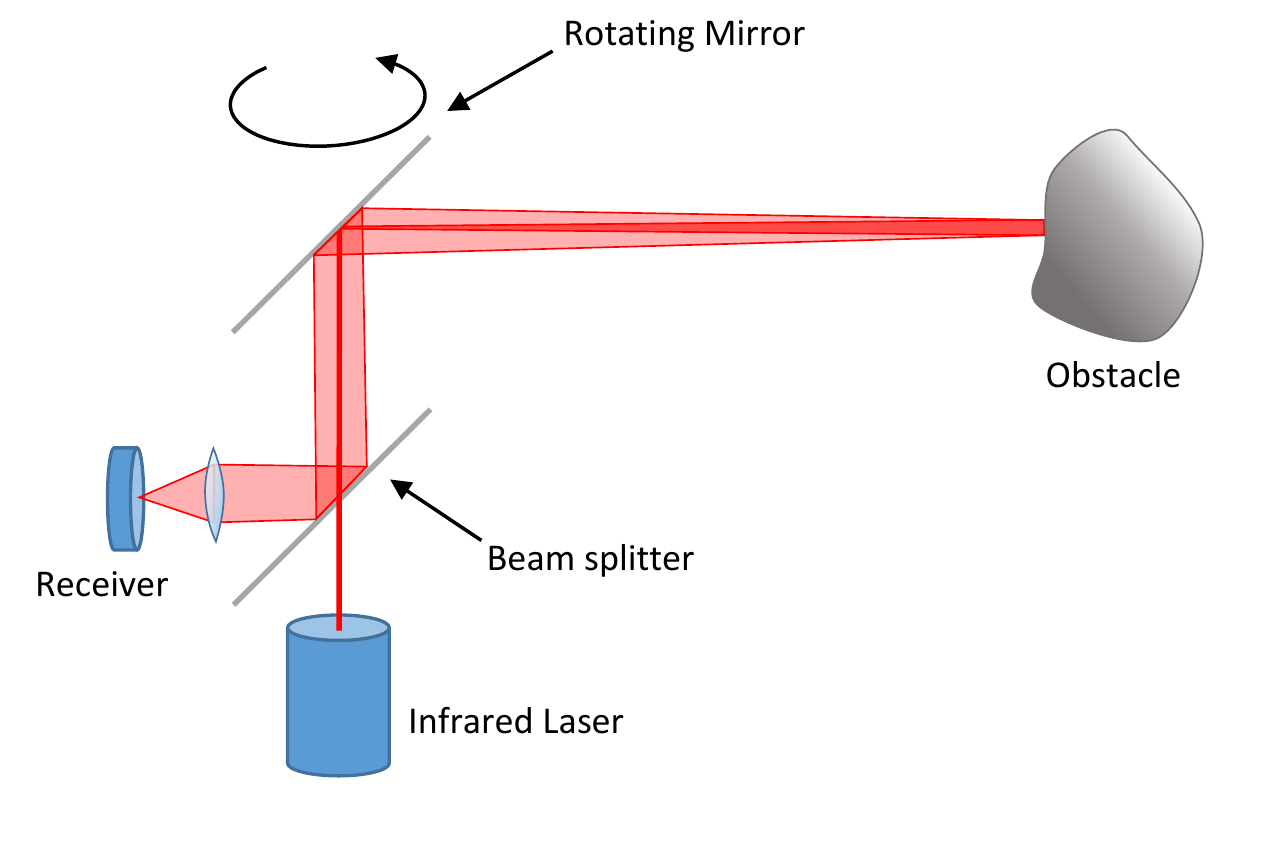}}		
		\caption[LiDAR working principle]{LiDAR working principle}		
		\label{fig:synoptiqueLidar}
	\end{figure}

\subsection{LiDAR point cloud distortions}

\begin{figure}[t]
\centering
	\subfloat[][Car trajectory during a LiDAR scan]{\includegraphics[width=0.43\columnwidth, trim = 65mm 135mm 65mm 55mm,clip]{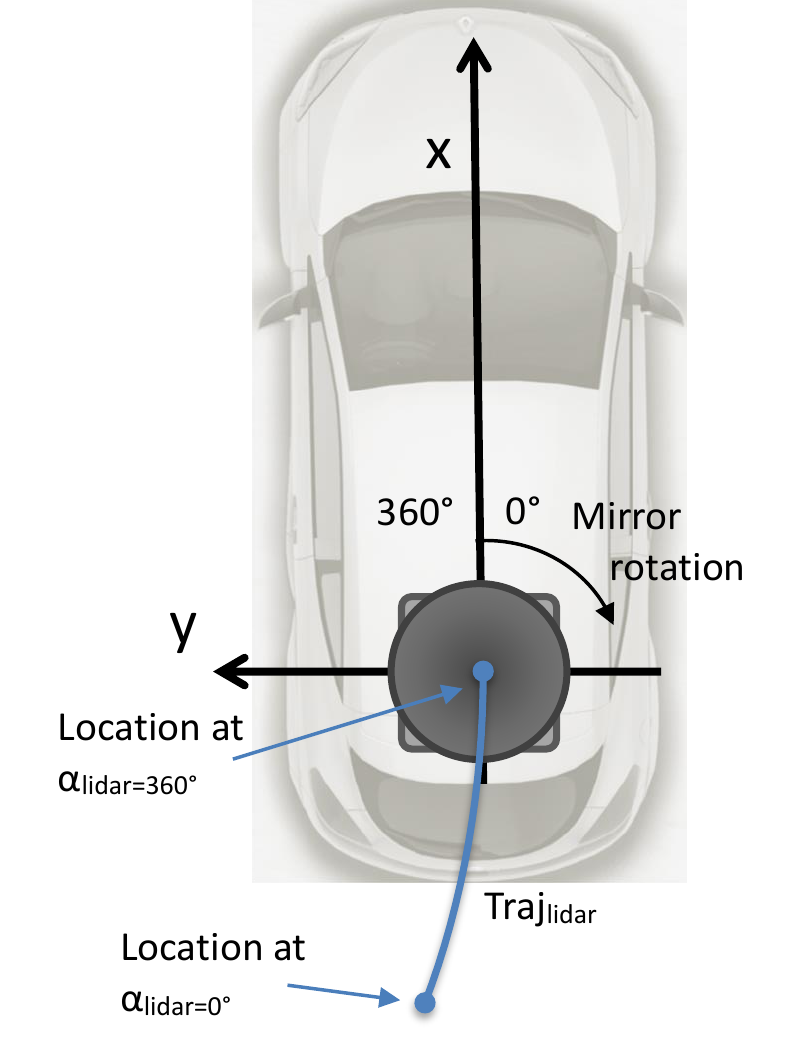}\label{fig:mvtVehicule}}$\qquad$
	\subfloat[][The car rotation is added to the laser rotation. The LiDAR viewpoint is not a single viewpoint anymore]{{\includegraphics[width=0.43\columnwidth, trim = 72mm 123mm 72mm 90mm,clip]{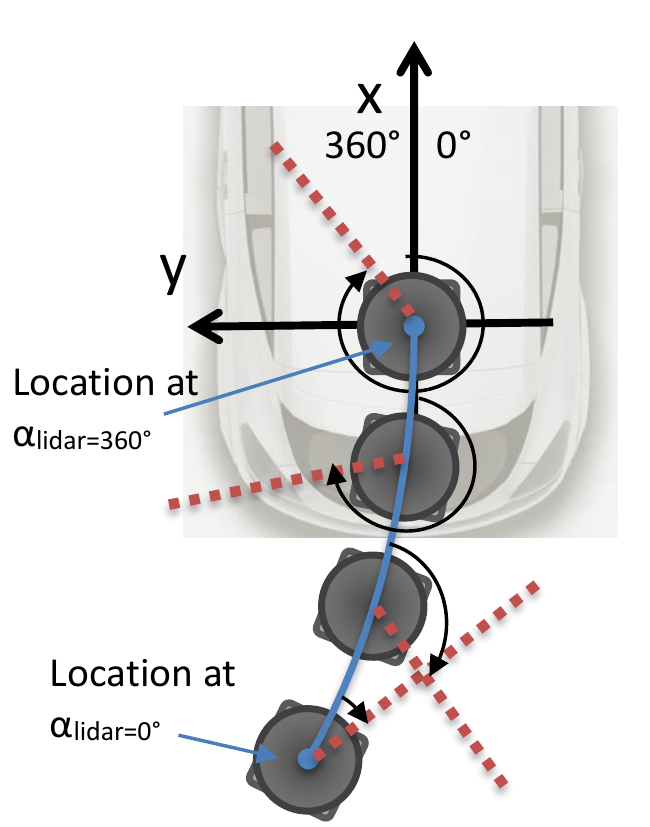}}\label{fig:mvtScan}}
	\caption{LiDAR scan distortions introduced by the car motion}
	\label{fig:trajectoireLidar}
\end{figure}

A LiDAR swipes the environment thanks to a mirror reflecting a laser (Figure~\ref{fig:synoptiqueLidar}). The mirror is rotated with a motor in order to measure a given field of view. A scan period  $T_s$ depends on the rotational speeds, which can vary from few Hertz to several hundred Hertz depending on the LiDAR model. During a LiDAR revolution, an mounted LiDAR can move (Figure~\ref{fig:mvtVehicule}), resulting in a non single viewpoint measurement. As it can be seen in Figure~\ref{fig:mvtScan}, the vehicle motion can be added to the mirror rotation causing the LiDAR measurement to be distorted.

Distortions are caused by the vehicle motion. They depend on the motion speed and the scan period $T_s$. For instance, let us consider a \emph{Velodyne HDL64} running at 10Hz:
\begin{itemize}
 \item  a linear motion at  50km/h causes a gap of 1.38m between the begin and the end of a scan.
 \item a rotational motion at 25$^{\circ}$/s creates a gap of 2.19m at 50m of the LiDAR.
\end{itemize}

Consequently, the 3D point cloud is distorted and might generate errors if the map is used for precise car location based on LiDAR only.

\begin{figure}[t]
\centering
{\includegraphics[clip=true,width=0.95\columnwidth]{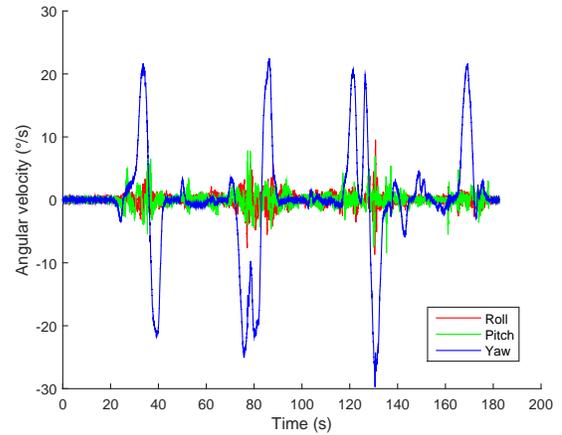}}
\caption{Angular velocities measured in the test set}
\label{fig:tauxRotation}
\end{figure}

\subsection{LiDAR point cloud correction}\label{sec:lpcc}
\begin{figure*}
\centering
\subfloat[][Overall view]{\includegraphics[clip=true,trim = 25mm 188mm 63mm 25mm,width=0.9\columnwidth]{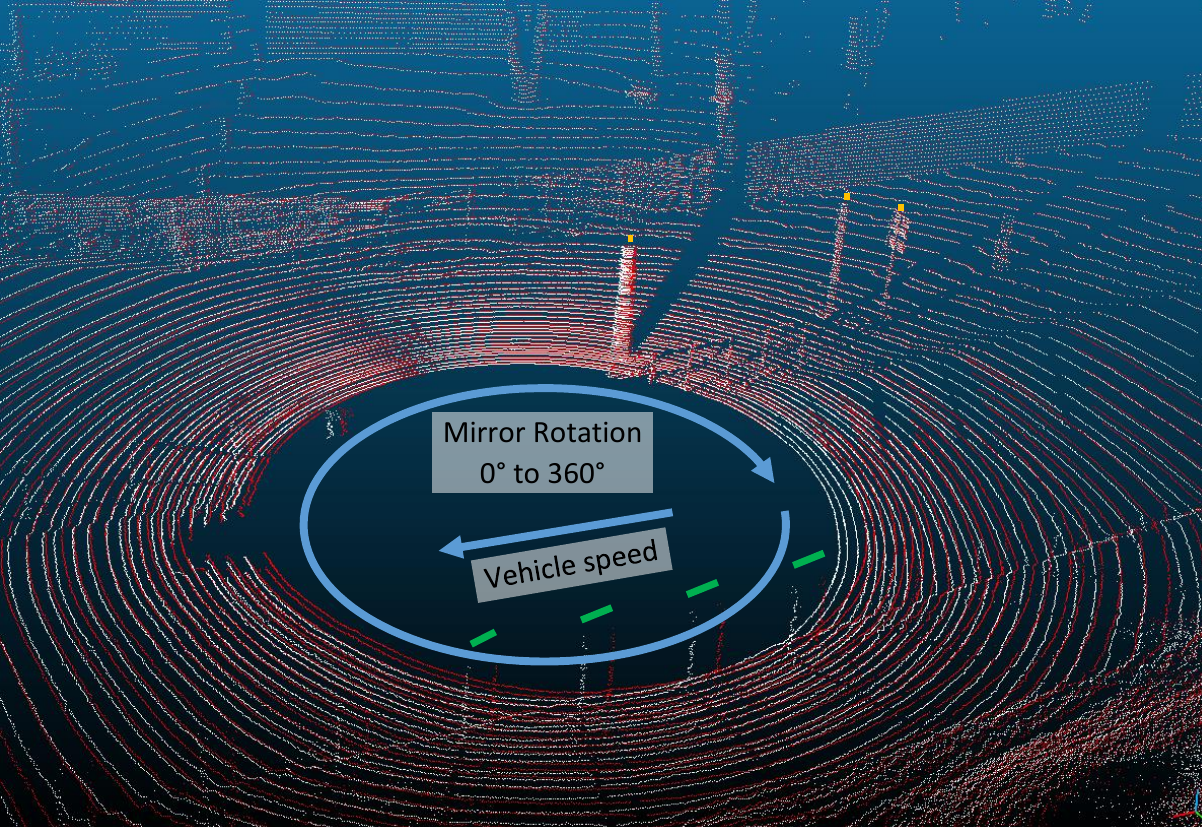}}~
\subfloat[][Close-up view]{\includegraphics[clip=true,trim = 70mm 191mm 95mm 75mm,width=0.9\columnwidth]{deformationLinear.pdf}}
\caption{Distortions in the LiDAR scan caused by a translation of 10m/s. Raw scan in white and corrected scan in red. Distortions are small at $\alpha_{LiDAR}$=360$^{\circ}$. Objects are close to each other. At $\alpha_{LiDAR}$=0$^{\circ}$, distortions are large. Foreground posts are 1m away from their ground truth locations (green lines).}
\label{fig:deformationLinear}
\end{figure*}

\begin{figure*}[t]
	\centering
	\subfloat[][The car and the LiDAR rotate clockwise. The LiDAR rotation is larger than 360$^{\circ}$. As it can be seen in the red box, the fence is duplicated]{\includegraphics[height=6cm]{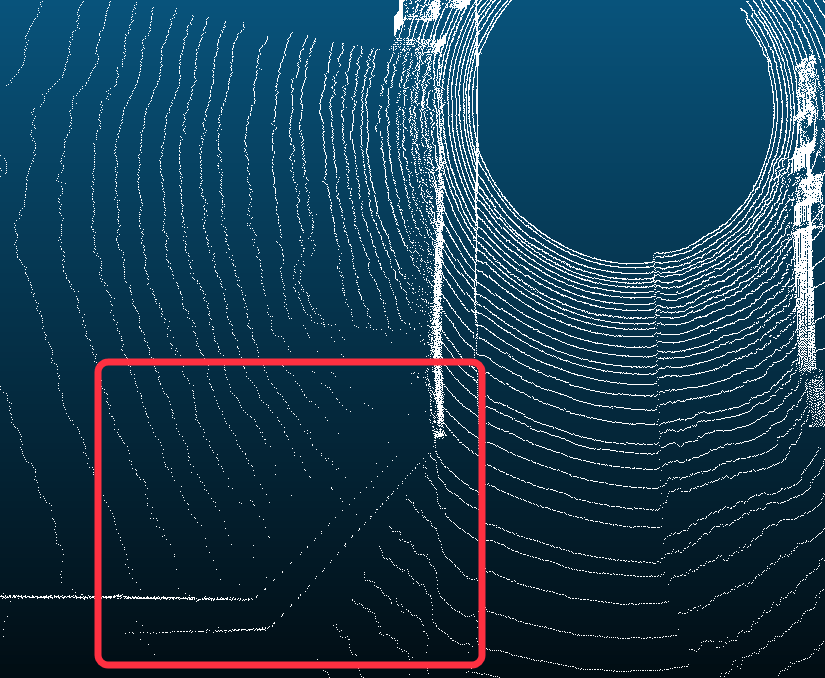}\label{fig:rotdeform}}~
	\subfloat[][Raw scan in white. Corrected scan in red. The fence is not duplicated anymore]{{\includegraphics[height=6cm]{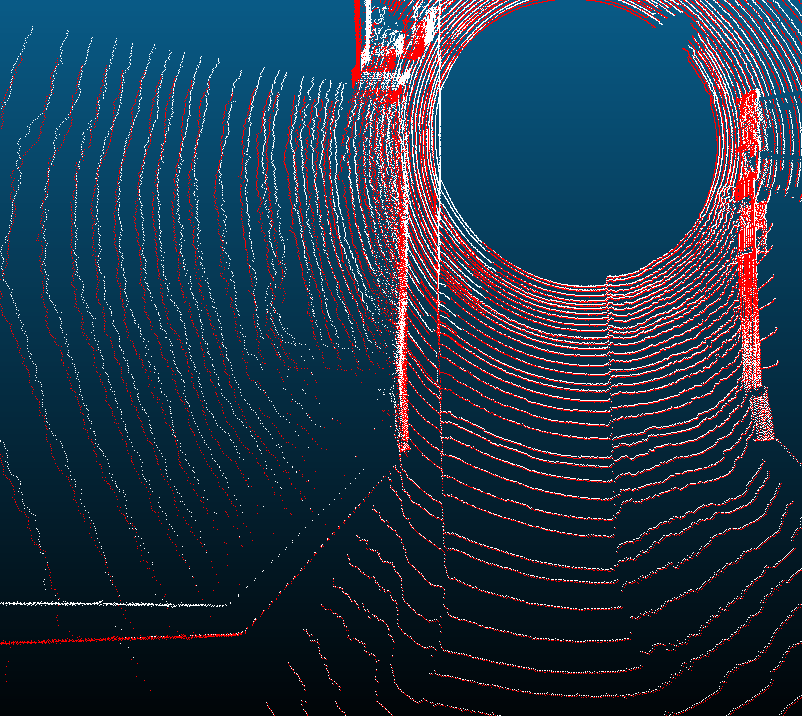}}\label{fig:rotCorrige}}
	\caption{Scan distortion for clockwise angular velocity of 25$^{\circ}$/s.}
	\label{fig:scanDeforme}
\end{figure*}

Short term road vehicle motion is mostly planar. Figure~\ref{fig:tauxRotation} highlights the angular velocities recorded in our test sets. It can be seen that the largest angular velocities are in the yaw rate component. Despite braking and acceleration stages, roll and pitch rates are rather small compared to the yaw rate. Consequently, a 2.5D correction can be performed.

Planar vehicle motion can be estimated from the car odometry. The linear motion $\Delta x$ and angular velocity $\Delta \theta$ can be estimated as follows:

	\begin{equation}
    			\label{eq:odometrieLineraire}
    			\Delta x = r\frac{\Delta\theta_{R}+\Delta\theta_{L}}{2}
	\end{equation}
 			\begin{equation}
    			\label{eq:odometrieRotation}
    			\Delta \theta = r\frac{\Delta\theta_{R}-\Delta\theta_{L}}{L}
	\end{equation}

	\begin{tabular}{rcl}
		$r$&:&wheel radius\\
		$L$&:&vehicle track\\
		$\Delta\theta_{R}$&:&angular position of the right wheel\\
		$\Delta\theta_{L}$&:&angular position of the left wheel\\
		$\Delta x$&:&linear motion\\
		$\Delta\theta$&:&angular velocity\\
	\end{tabular}
\ \\

Several methods exist to estimate the car trajectory from vehicular motion estimations. \emph{Runge-Kutta} method can be regarded as the most computationally efficient and stable method. It integrates vehicular motion recursively as follows:
			
	\begin{equation}	    		    	
    	\begin{bmatrix} X \\ Y \\ \theta \\ \end{bmatrix}_i =\begin{bmatrix} X \\ Y \\ \theta \\ \end{bmatrix}_{i-1} + \begin{bmatrix} \Delta x_i \cos(\theta_{i-1} + \Delta\theta_i/2) \\ \Delta x_i \sin(\theta_{i-1} + \Delta\theta_i/2) \\ \Delta\theta_i \\ \end{bmatrix}_{i}
    	\label{eq:odoVecteurEtat}
	\end{equation}

	\begin{tabular}{rcl}
		$X, Y$&:&planar location\\
		$\theta$&:&heading\\
		$\Delta x_i$&:&linear motion at time $i$\\
		$\Delta\theta_i$&:&angular motion at time $i$\\
	\end{tabular}

\ \\

When the LiDAR does not move while scanning, the LiDAR measurements can estimated as follows:

\begin{equation}
\label{eq:spheriqueCartesien}
E_b=\begin{bmatrix} X_b \\ Y_b \\ Z_b \\ \end{bmatrix}=\begin{bmatrix} cos(\omega)cos(\alpha)  \\ cos(\omega)sin(\alpha) \\ sin(\omega) \\ \end{bmatrix}*d
\end{equation}

	\begin{tabular}{rcl}
		$\omega$&:&LiDAR ray inclination\\
		$\alpha$&:&LiDAR ray azimuth\\
		$d$&:&measured distance\\
		$X_b, Y_b, Z_b$&:&LiDAR point location in the LiDAR frame\\
	\end{tabular}

\ \\

When the LiDAR moves while the scanning period $T_s$, Equ.~\ref{eq:odoVecteurEtat} can be used to estimate the LiDAR location at any given measurement. As we consider planar motion, the LiDAR correction only depends on the LiDAR azimuth $\alpha$.

It is better to obtain the corrected LiDAR scan when $\alpha_{lidar}=360^{\circ}$ than when $\alpha_{lidar}=0^{\circ}$ (figure~\ref{fig:mvtVehicule}). In fact, the last frame is fixed with respect to the vehicle and does not depend on the vehicle speed. Consequently, the LiDAR location $P_\alpha$ must be back projected with respect to the LiDAR azimuth $\alpha$ and the vehicle odometry:

	\begin{align}	
		\Delta x_\alpha &= -\Delta x\frac{\alpha}{2\pi}\nonumber \\    		    	
		\Delta \theta_\alpha &= -\Delta \theta\frac{\alpha}{2\pi} \nonumber \\
    	P_{\alpha}&=\begin{bmatrix} X_\alpha \\ Y_\alpha \\ \theta_\alpha \\ \end{bmatrix} = \begin{bmatrix} -\Delta x_\alpha \cos(\theta_\alpha  -\Delta\theta_\alpha/2) \\ -\Delta x_\alpha \sin(\theta_\alpha  -\Delta\theta_\alpha/2) \\ -\Delta\theta_\alpha \\ \end{bmatrix}
    	\label{eq:odoNegativeVecteurEtat}
	\end{align}

\begin{tabularx}{\columnwidth}{rcX}
	$\theta_\alpha$&:&LiDAR ray heading when the LiDAR ray azimuth is $\alpha$\\
\end{tabularx}

\ \\
Once the LiDAR location has been estimated, we can correct Equation~\ref{eq:spheriqueCartesien} in order to take into account the LiDAR frame motion:

	\begin{align}	
	E_c&=\begin{bmatrix} X_c \\ Y_c \\ Z_c \\ \end{bmatrix}=\begin{bmatrix} X_\alpha \\ Y_\alpha \\ 0 \\ \end{bmatrix} \nonumber\\
&+\begin{bmatrix} cos(\theta_\alpha) & -sin(\theta_\alpha) & 0  \\ sin(\theta_\alpha) & cos(\theta_\alpha)&0\\ 0 & 0 & 1 \end{bmatrix} \begin{bmatrix} cos(\omega)cos(\alpha)  \\ cos(\omega)sin(\alpha) \\ sin(\omega) \\ \end{bmatrix}*d
    	\label{eq:correctionScan}
	\end{align}

Transformation matrices must be updated at any new $\alpha$ while the LiDAR is scanning.


\subsection{Singular motion distortions}

Two main types of singular vehicular motion exist : translations and rotations. They do not cause the same distortions:

\begin{itemize}
\item Linear motion (Figure~\ref{fig:deformationLinear}) : LiDAR points are projected at the end of the LiDAR scan, i.e. $\alpha_{lidar}=360^{\circ}$. While they are larger when considering the beginning of the scan, distortions are smaller when choosing the reference frame at the end of the scan. Given such vehicle motion and LiDAR mirror rotation, scans are more distorted on the right-hand side of the vehicle than the left-hand side.
\item Vehicle motion travelling around a curve  : LiDAR mirror rotation is added to the vehicle rotation (Figure~\ref{fig:scanDeforme}). On the one hand, as it can be seen in Figure~\ref{fig:mvtScan}, depending on the rotation direction, LiDAR scan measurements can cover more the 360$^{\circ}$ in the scene. Objects close to the beginning of the scan can be measured twice. On the other hand, less than 360$^{\circ}$ of the scene can be measured and objects might be missing. Missing objects will not be recovered by scan correction. However, measured points will be corrected.
\end{itemize}

Vehicle trajectories encompassing these two types of motions includes variants of both types of distortions.

\subsection{3D reconstruction}

In order to assess the 3D reconstruction quality, we used 3DTK tool\footnote{\url{http://slam6d.sourceforge.net/}}. It includes two stages:
\begin{itemize}
\item First, consecutive scans are matched. The matching process~\cite{nuchter2007p} is optimized with ICP (\emph{Iterative Closest Point} \cite{besl1992a}).
\item Secondly, SLAM-based loop closure is performed in order to globally optimize the 3D point cloud~\cite{sprickerhof2009e}.
\end{itemize}

The tool requires 3D point cloud of every scan and the corresponding LiDAR poses when the scan is outputted by the LiDAR. Poses are used as the initialization of the ICP process.

\section{EXPERIMENTS}

\begin{figure*}[t]
	\centering
	\subfloat[][Measurement setup]{\includegraphics[width=0.9\columnwidth]{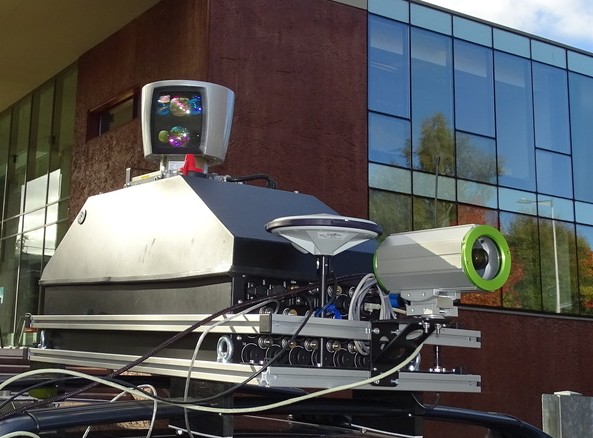}}~~~~~
	\subfloat[][Sensor locations]{{\includegraphics[trim = 15mm 155mm 15mm 25mm, clip,width=0.9\columnwidth]{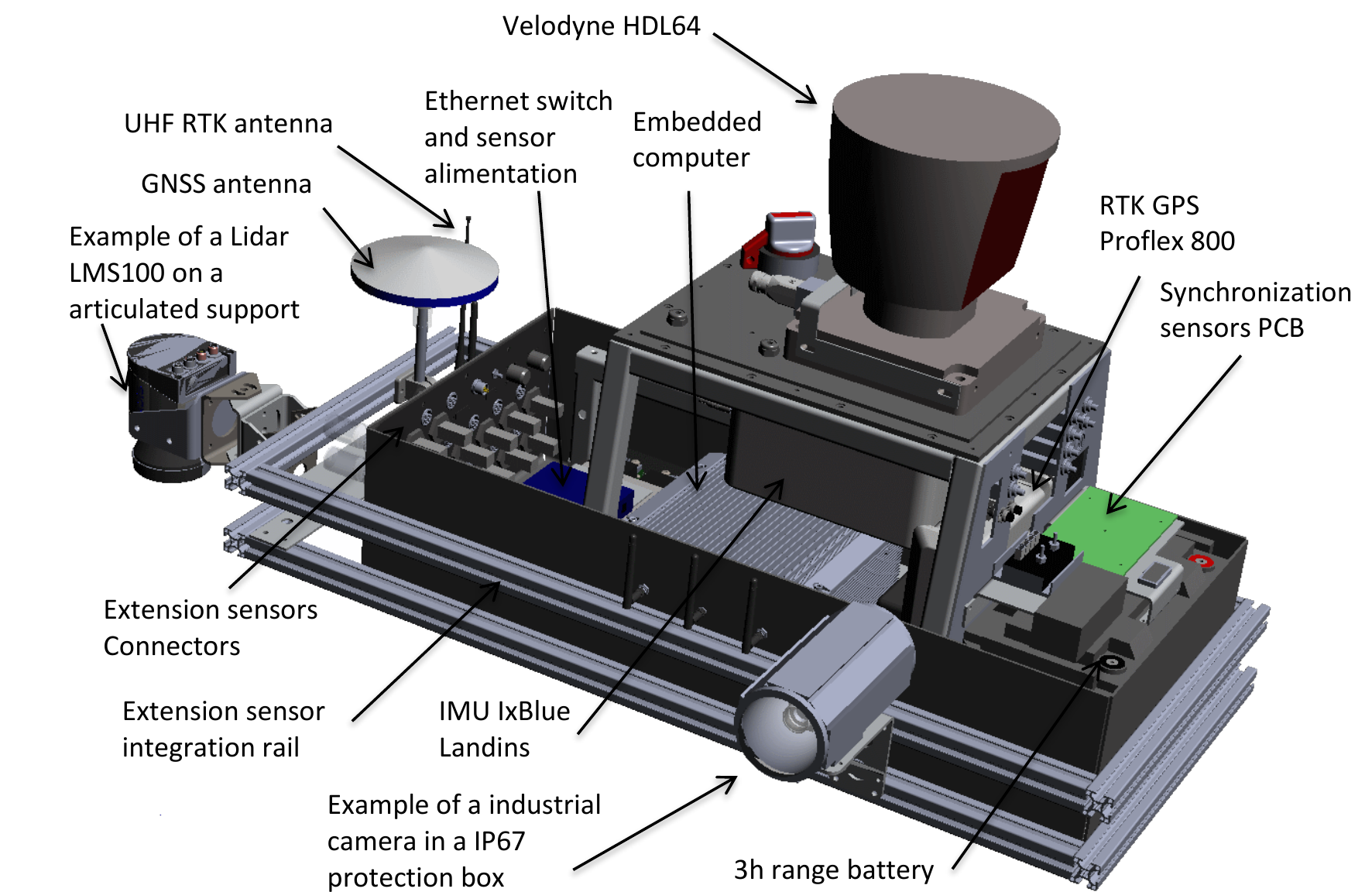}}}
	\caption{Ground truth setup used during our experiments. It includes a centimeter grade positioning device and a 64 layer LiDAR Velodyne HDL64 .}
\label{fig:coffreDeToit}
\end{figure*}

The ground truth setup is mounted on a car as a roof box (Figure~\ref{fig:coffreDeToit}). It includes the following devices:
\begin{itemize}
\item IMU \emph{FOG} : Landins IxBlue
\item Differential RTK GPS Proflex 800
\item Peiseler odometer mounted on the rear wheel
\item Multi-layer LiDAR Velodyne HDL-64e
\item i7-3610 embedded computer equipped with a SSD of 1 To
\end{itemize}

The car location accuracy is given in Table ~\ref{tab:landinsPrecision}:

\begin{table}
\centering
\begin{tabular}{lcc}
\hline
\textbf{Mode} & \textbf{GPS RTK} & \textbf{60s without GPS}\\
\hline
\hline
True Heading ($^{\circ}$) & 0.01 & 0.01\\
Roll/Pitch ($^{\circ}$) & 0.005 & 0.005\\
Location X,Y (m) & 0.02 & 0.1\\
Locattion Z (m) & 0.05 & 0.07\\
\hline
\end{tabular}
\caption{Accuracy of the LiDAR location}
\label{tab:landinsPrecision}
\end{table}

Linear displacement $\Delta x$ are obtained from ABS sensors on the CAN bus of the vehicle. Angular displacements $\Delta \theta$ are measured by the ESP gyroscope and also read on the CAN bus.

The test set was recorded in a suburban area. It encompasses different types of movement including mixes of translations and rotations.

The IMU outputs the car position in the WG84 frame. Positions are converted in Universal Transverse Mercator (UTM) in order to use a Cartesian reference frame. The IMU information are not used to estimate the car motion during a scan. GPS data are required by the 3DTK tool as SLAM pose initial estimates.

RTMaps framework\footnote{\url{https://intempora.com/products/rtmaps.html}} was used to obtain synchronized data from the ground truth setup and the CAN bus. A specific component was developed to post process the LiDAR scan given the LiDAR poses estimated by the approach suggested in Section~\ref{sec:lpcc}.

\section{RESULTS}

As it can be seen in Figure~\ref{fig:deformationLinear}, the car motion is mainly linear. As expected, the gap between the corrected point cloud and the raw point cloud is roughly the distance travelled during time $T_s$. As it can be seen in Figure~\ref{fig:scanDeforme}, movements with large angular velocities introduce duplicates in the measured scene. The 3DTK tool does not include scan distortion removal. It can be clearly seen that the fence is duplicated when raw data are considered The duplicated fence is removed from the corrected scan with CAN-bas data prior to scan registration in the SLAM framework (c.f. Figure~\ref{fig:rotCorrige}).

This evaluation is qualitative. In order to assess the performance from a quantitative standpoint, we propose to use the following metrics :

\begin{itemize}
\item Mean Point-to-Point error during the ICP process.  ICP scan matching should be more challenging when the scans are distorted  resulting in larger matching distances.
\item Compactness of the 3D map. Raw scans may be more spread than corrected scans. Consequently, their representation should be less compact than corrected scans.
\end{itemize}

	\begin{figure}
	\centering
	    {\includegraphics[width=0.8\linewidth, clip=true]{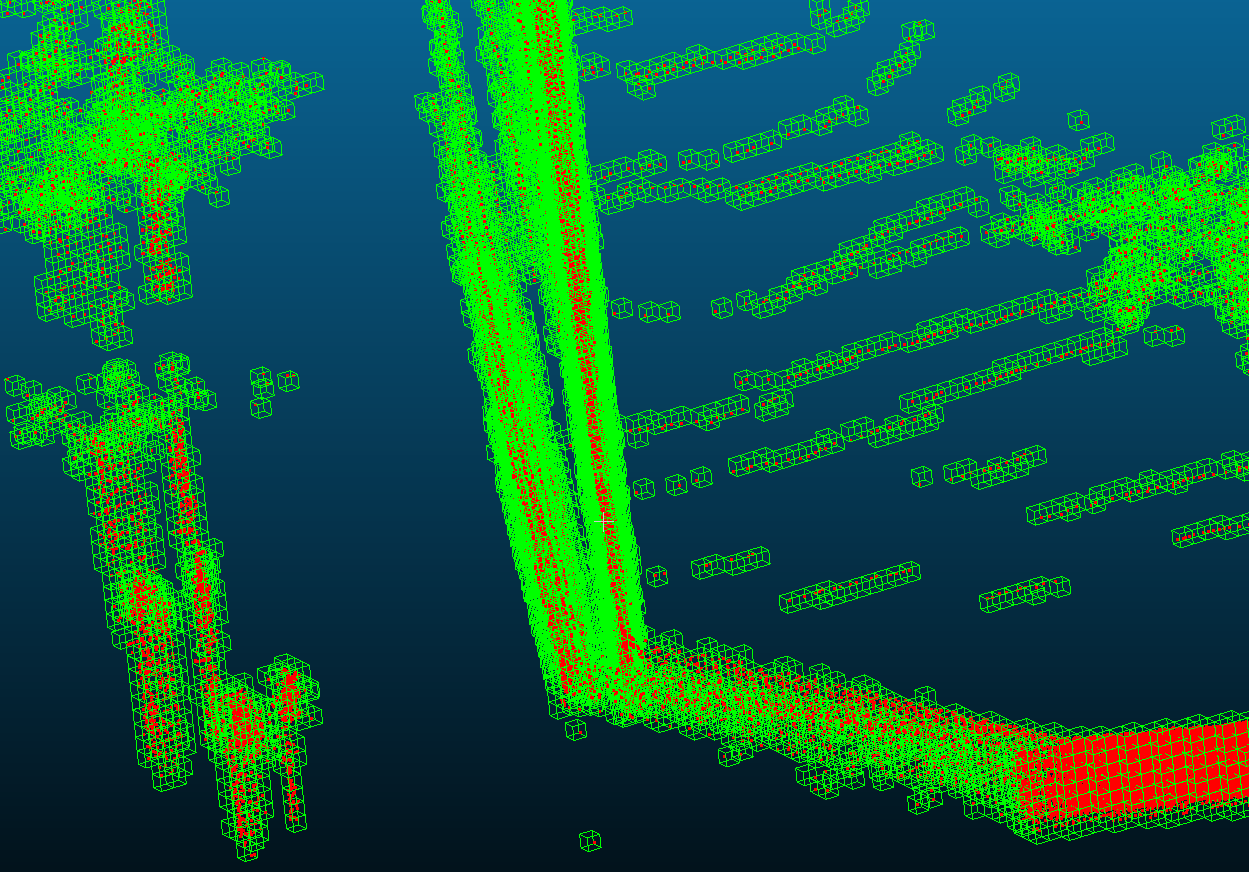}}		
		\caption{3D reconstruction of the clockwise rotation. 0.2m octree cells are shown in green~\ref{tab:octree}. The duplicated fence (c.f. Figure~\ref{fig:rotdeform}) must be stored in the octree while the corrected scan would not required these extra cells.}		
		\label{fig:octree}
	\end{figure}

\begin{figure*}[t]
	\centering
	\subfloat[][58 scans during a linear motion at 10m/s.]{{\includegraphics[clip=true,trim = 5mm 2mm 10mm 5mm,width=0.9\columnwidth]{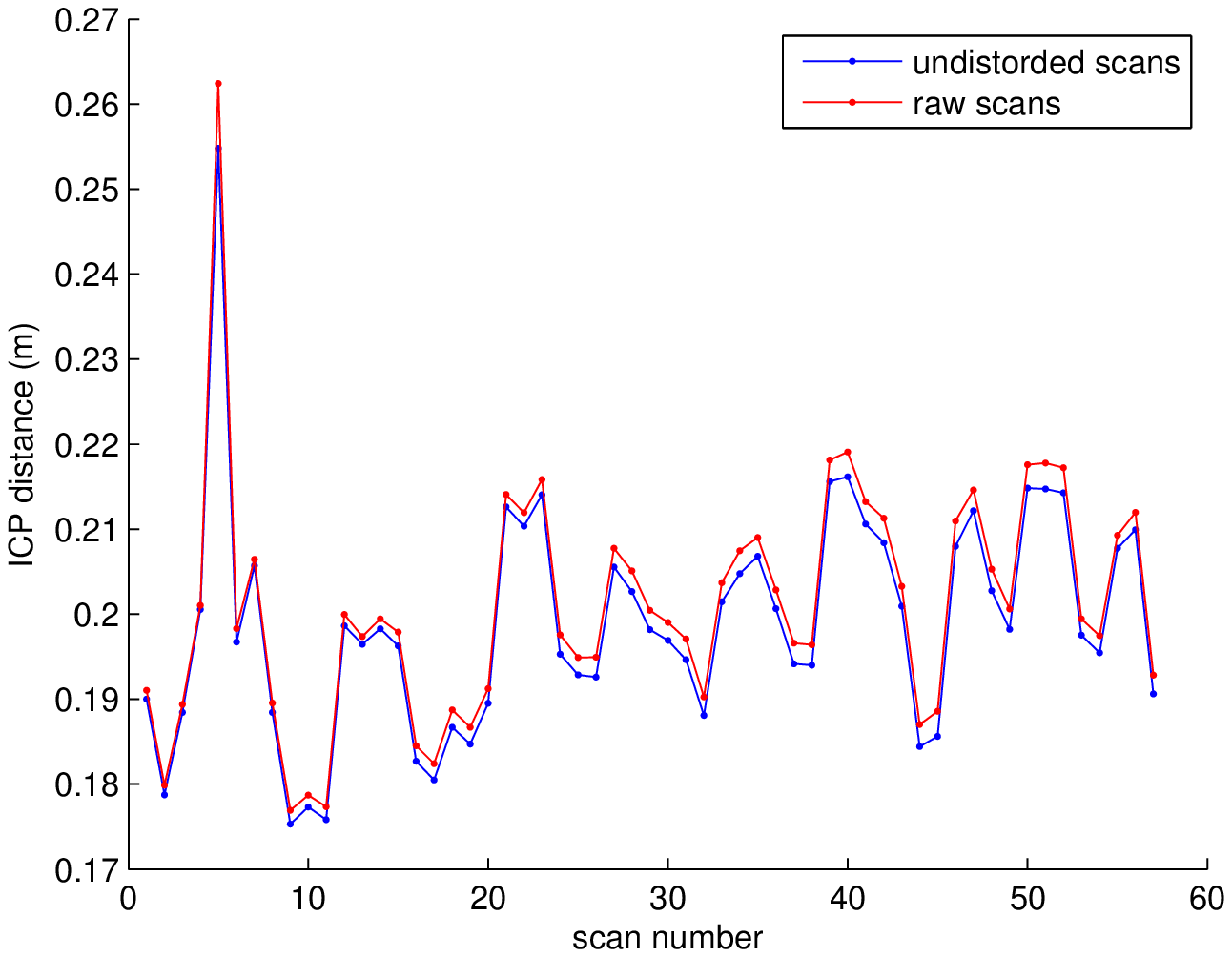}}}~~
	\subfloat[][15 scans during a rotation at 25$^{\circ}$/s.]{{\includegraphics[clip=true,trim = 5mm 2mm 10mm 5mm,width=0.9\columnwidth]{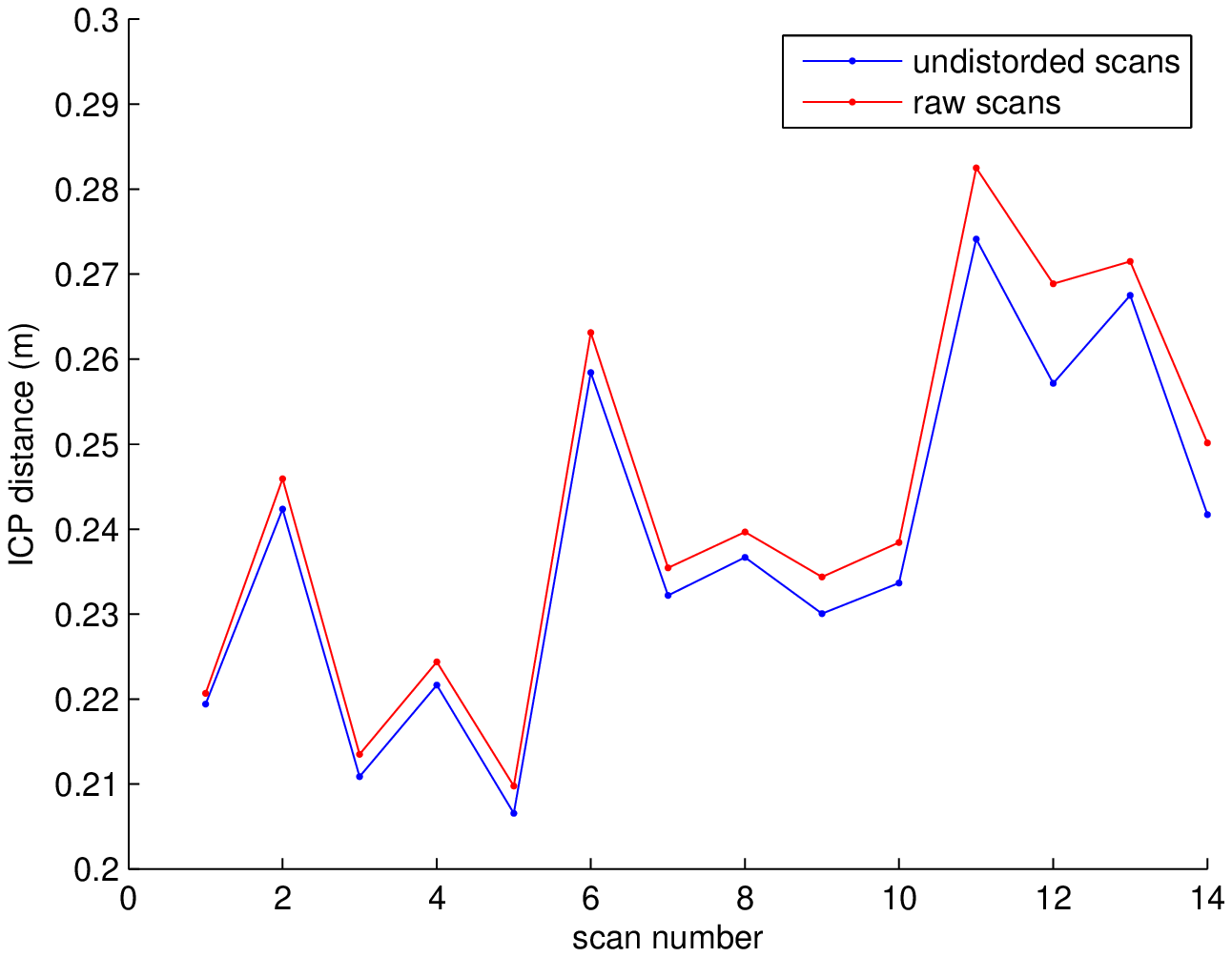}}\label{fig:icpLineaire}}
	\caption{Matching distance given consecutive scan with ICP.}
	\label{fig:resultatsICP}
\end{figure*}

These metrics were used with two singular movements : a translation and a rotation. The ICP matching score is computed from two consecutive scans. The scan distances are shown in Figure~\ref{fig:resultatsICP}. The difference between raw scans and corrected scan is rather small given the metrics used. As a matter of fact, only a subset of the points suffers from distortions. Moreover, matching two distorted scans still include a lot of common points especially when large horizontal plane exists in the scene. Consequently, despite errors exist in the scans, they do not affect ICP significantly. However, the map itself is corrupted. The error between raw and corrected scans is even smaller for linear motion (c.f. Figure~\ref{fig:icpLineaire}). While the impact of car motion does not seem large in our scenario, it would be larger if the road was traveled in opposite directions. Distortions would be on the right-hand side of the road on one way and left-hand side on the other way.

%

\begin{figure}
{\includegraphics[clip=true,width=0.95\columnwidth]{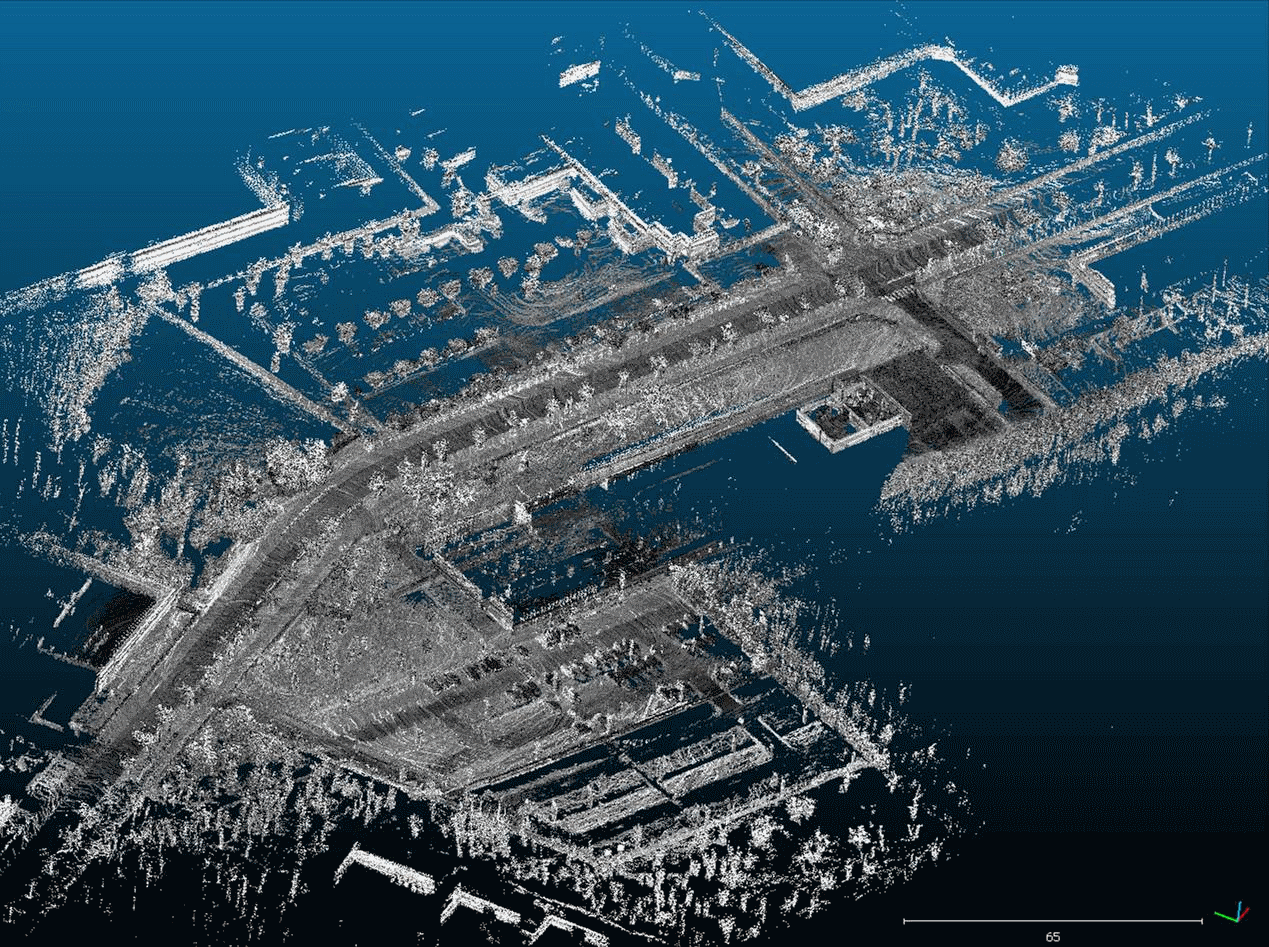}}
\caption{3D reconstruction from 225 scans of a \emph{Velodyne HDL64}. The grey scale corresponds to the LiDAR remission measurement. The 3D map corresponds to $408m\times275m\times50m$.}
\label{fig:ciseNuageDePoints}
\end{figure}

\begin{figure}
\centering
\subfloat[][Raw scan fusion]{\includegraphics[width=1\columnwidth]{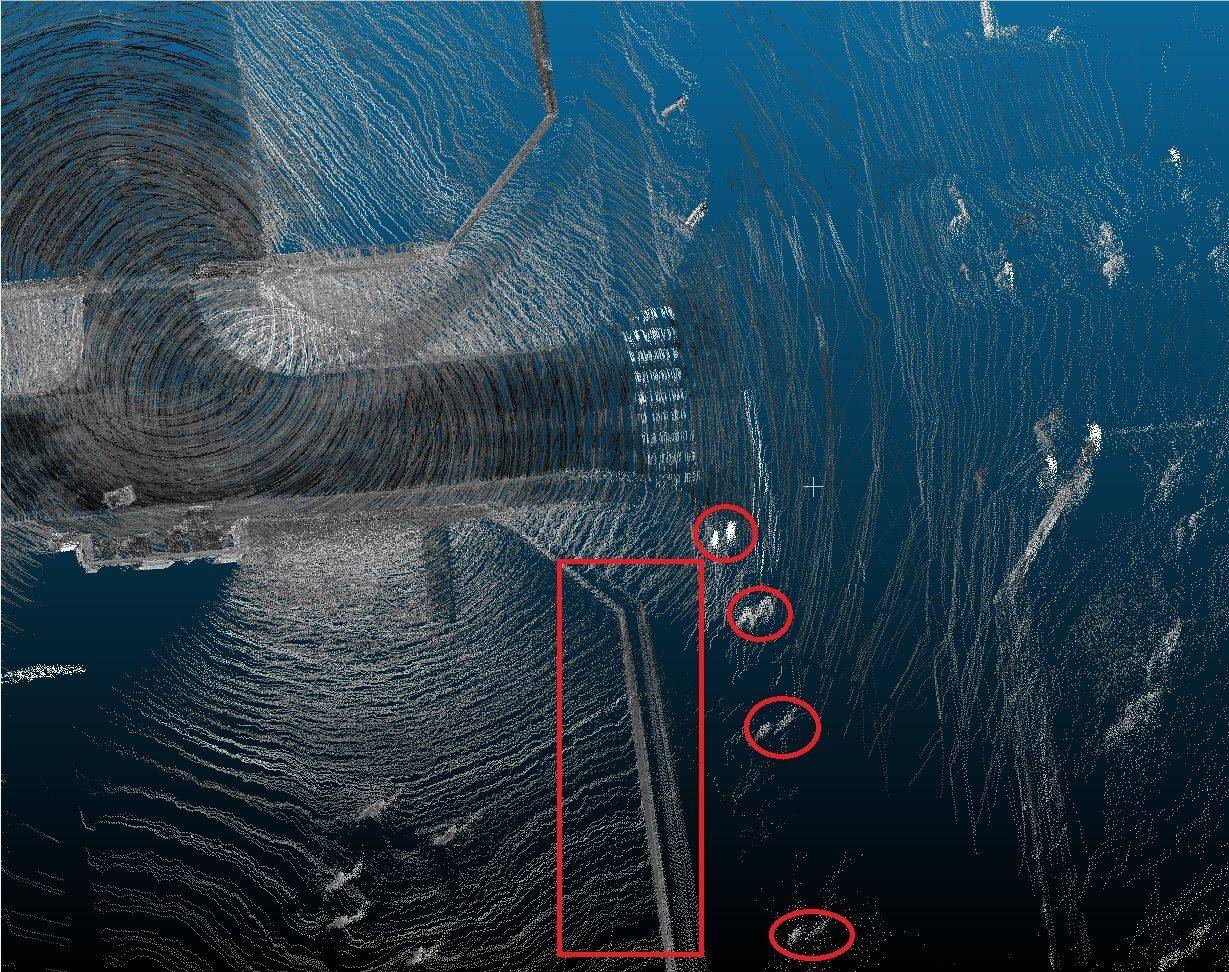}}\\
\subfloat[][Corrected scan fusion]{\includegraphics[width=1\columnwidth]{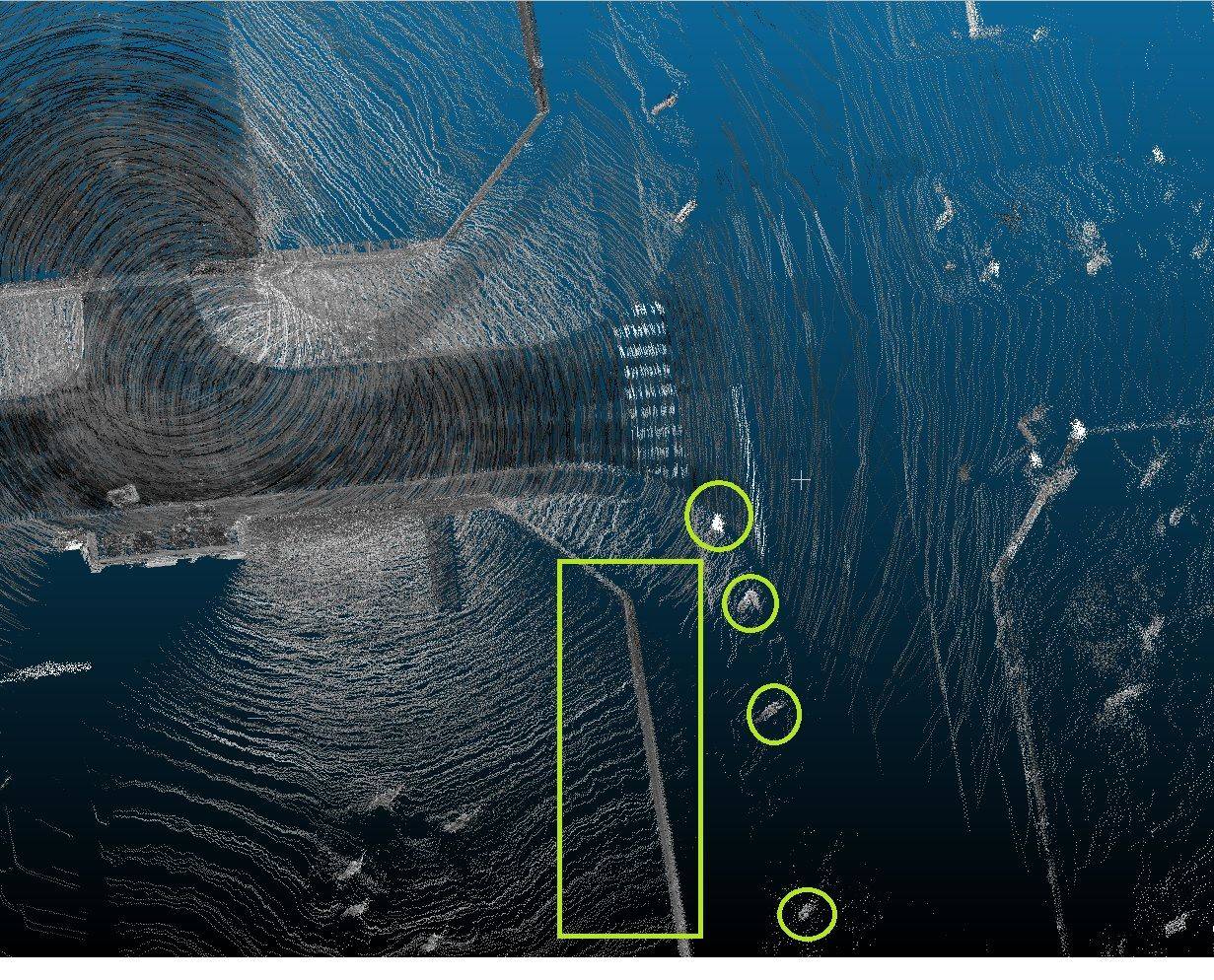}}
\caption{Close-up view of the large-scale 3D map\\\emph{The fence and trees are duplicated in the map build from raw scans}}
\label{fig:issuerotate}
\end{figure}

Given both motion types, we stored the 3D point cloud in an octree (c.f. Figure~\ref{fig:octree}). The octree cells were set to 0.1m. The number of occupied cells allows to measure the occupied space. Table~\ref{tab:octree} gives the number of occupied cells given different motion and scan types. As mentioned earlier, only a subset of the 3D points are distorted. As we can see in Figures~\ref{fig:rotdeform} and~\ref{fig:octree}, only the duplicated fence seems to impact the number of occupied cells.

\begin{table}[t]
\caption{3D reconstruction modeling with an octree : number of $0.1\times 0.1m$ occupied cells.}
\centering

\begin{tabular}{|c|c|c|}
\hline
Motion type & Raw scans & Corrected scans \\
\hline
\hline
Linear (10m/s) & 1918170 & 1894744 \\
\hline
Rotation (25$^{\circ}$/s) & 654488 & 639094 \\
\hline
\end{tabular}

\label{tab:octree}
\end{table}

In order to qualitatively validate our work, we reconstructed a 3D map of our test set from corrected scans(c.f. Figure~\ref{fig:ciseNuageDePoints}). It uses 225 LiDAR scans and encompasses $408m\times275m\times50m$. The 3D point cloud quality is fair: lines are straights, posts are correctly aligned. Figure~\ref{fig:issuerotate} gives a close-up view on a given part of the large-scale 3D map. It can be seen that when raw scan are used the fence is duplicated and fuzzy despite the fusion of several scans. The same problem arises with posts and trees.

\section{CONCLUSION}

In this paper, we have studied the impact of car mouvements on LiDAR scans. High speed linear motions and slow speed by large angular velocity introduce the larger distortions in the LiDAR scans. Objects may be duplicated or missing from the resulting large-scale 3D map.The LiDAR should be carefully oriented to place the resulting blind spot away from region of interests for the control of the car. Our scan correction approach is simple and based on the estimation of the 2.5D motion of the vehicle carrying the LiDAR.

The metrics used do not highlight enough the improvements resulting from scan correction. Consequently, future works while be focused on scanning the environnement with \emph{Leica C10}\footnote{\url{http://www.leica-geosystems.fr/fr/Leica-ScanStation-C10_79411.htm}} to obtain the ground truth. As a result, we will be able to obtain the point-to-point distance between the ground truth, the raw scan 3D map and the corrected scan 3D map

\bibliographystyle{IEEEtran}
\bibliography{IEEEabrv,biblioRFIA}

\end{document}